\documentclass[10pt,twocolumn,letterpaper]{article}

\usepackage{iccv}
\usepackage{times}
\usepackage{epsfig}
\usepackage{graphicx}
\usepackage{amsmath}
\usepackage{amssymb}
\usepackage{cite}
\usepackage{amsfonts}
\usepackage{algorithmic}
\usepackage{textcomp}
\usepackage{xcolor}
\usepackage{algorithm}
\usepackage{marvosym}
\usepackage{multirow}
\usepackage{multicol}

\usepackage[pagebackref=true,breaklinks=true,letterpaper=true,colorlinks,bookmarks=false]{hyperref}

\iccvfinalcopy 

\def\iccvPaperID{154} 

\ificcvfinal\fi
\begin{document}

\title{Regularizing Neural Networks for Future Trajectory Prediction via Inverse Reinforcement Learning Framework}

\author{Dooseop Choi, Kyoungwook Min, Jeongdan Choi\\
ETRI\\
Republic of Korea\\
{\tt\small d1024.choi@etri.re.kr}
}

\maketitle

\begin{abstract}
   Predicting distant future trajectories of agents in a dynamic scene is not an easy problem because the future trajectory of an agent is affected by not only his/her past trajectory but also the scene contexts. To tackle this problem, we propose a model based on recurrent neural networks (RNNs) and a novel method for training the model. The proposed model is based on an encoder-decoder architecture where the encoder encodes inputs (past trajectories and scene context information) while the decoder produces a trajectory from the context vector given by the encoder. We train the networks of the proposed model to produce a future trajectory, which is the closest to the true trajectory, while maximizing a reward from a reward function. The reward function is also trained at the same time to maximize the margin between the rewards from the ground-truth trajectory and its estimate. The reward function plays the role of a regularizer for the proposed model so the trained networks are able to better utilize the scene context information for the prediction task. We evaluated the proposed model on several public datasets. Experimental results show that the prediction performance of the proposed model is much improved by the regularization, which outperforms the-state-of-the-arts in terms of accuracy. The implementation codes are available at https://github.com/d1024choi/traj-pred-irl/. 
\end{abstract}

\section{Introduction}

\begin{figure}
\centerline {\includegraphics[width=8.0cm]{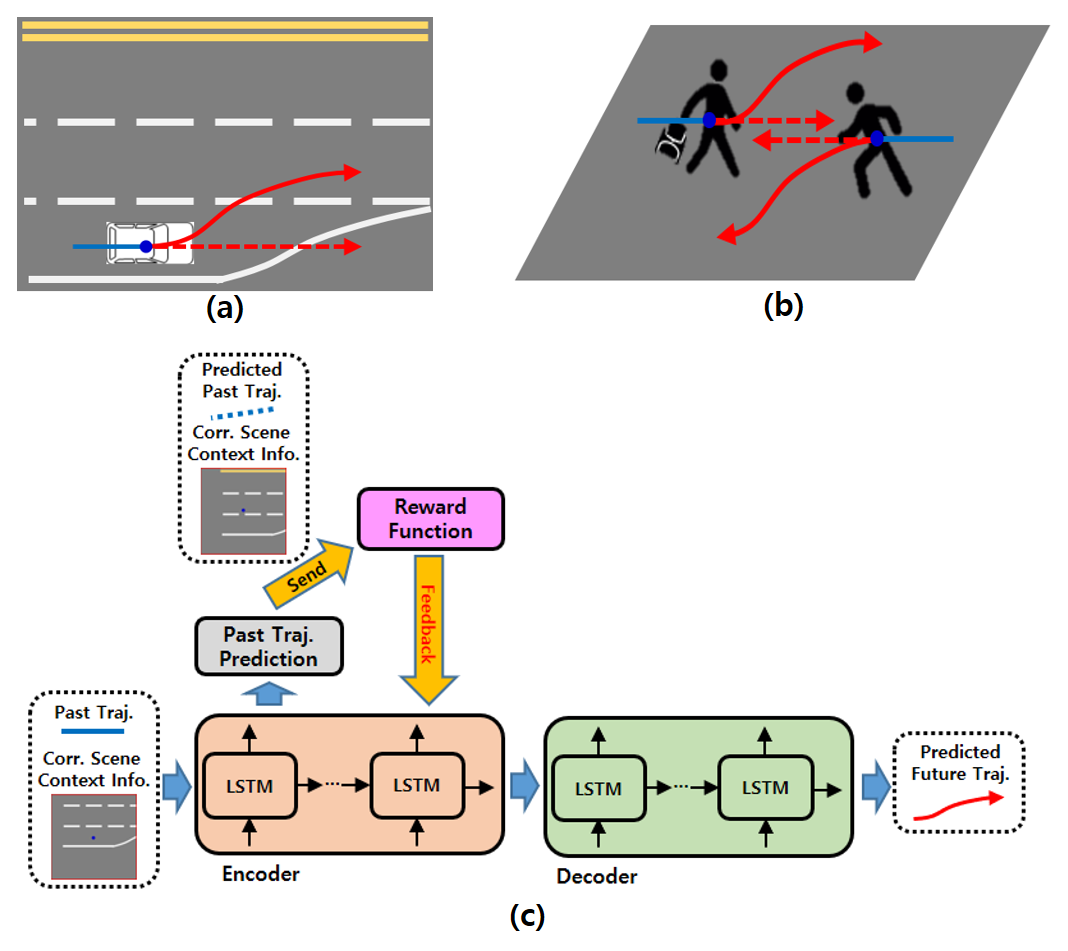}}
\caption{The static scene context (a) and dynamic scene context (b) are good clues for predicting the future movement of agents in a dynamic scene. (c) The conceptual explanation of the proposed approach. The encoder produces informative vectors by encoding the input data while the decoder predicts the future trajectory from the last informative vector. The informative vectors are also utilized to predict the positions in the past trajectory. The reward function evaluates the positions predicted by the encoder along with the scene context information caused by the positions and is simultaneously trained with the encoder and decoder. As a result, the encoder learns how to extract meaningful features from the inputs and encode the features efficiently, which in turn helps the decoder predict the future trajectory more accurately.}
\label{fig:one}
\end{figure}

The future trajectory prediction of agents in a dynamic scene has long been a great interest in many research fields such as autonomous driving, video surveillance. Predicting the distant future trajectory, however, is not an easy task because the target position of an agent is not known in advance. In addition, the motion of an agent is affected by not only his/her past motion but also the scene contexts such as the motion of the neighboring agents (dynamic scene context) and road structure (static scene context). Figure~\ref{fig:one} shows examples.   

Recently, many approaches have been proposed aiming at further improving the prediction accuracy by utilizing the scene context information. $\textit{Social pooling}$~\cite{alahi2, Lee, Xue, Gupta, Sadeghian3} and $\textit{attention mechanism}$ ~\cite{SadeghianA, Xu} are the representative techniques where the former deals with the interactions between multiple agents while the later mainly deals with the interaction between an agent and the static scene.

While existing approaches have paid attention to how to use encoded features from the scene context information, we, in this paper, will focus on how to make neural networks extract meaningful features from the scene context information and encode the features efficiently so that the networks can predict the future trajectory more accurately. To this end, we propose a model for the prediction task and its training method. The proposed model is composed of two modules, $\textit{trajectory prediction module}$ (TPM) and $\textit{reward function}$ (RF) as shown in Figure~\ref{fig:one}. TPM is based on an encoder-decoder RNN architecture, where the encoder encodes inputs (past trajectory and scene context information) while the decoder predicts the future trajectory from the context vector given by the encoder. RF produces a reward value for visiting a specific state (a position and the scene context information caused by the position). RF is introduced to regularize the encoder, hoping that the encoder learns how to extract meaningful features from the inputs and encode the features efficiently for the prediction task.

The main contribution of this paper can be summarized as follows: (1) an encoder-decoder RNN architecture based model, which produces a future trajectory from the past trajectory and the scene context information, is proposed, (2) maximum margin inverse reinforcement learning framework~\cite{Abbeel} is introduced to regularize the proposed model, (3) an alternation method for the simultaneous training of TPM and RF is proposed.

\section{Related Works}
\textbf{Agent-agent interactions} Early studies usually focused on building mathematical models~\cite{Helbing, Lerner, Pellegrini, Treuille, Antonini, Wang, Tay, Yi, Robicquet, Pellegrini2, Scovanner, Mehran} or hand-crafted features~\cite{Rodriguez, alahi, Morris2, Wang3} to better represent the interactions between neighboring agents. Recently, thanks to the vast amount of training samples available, RNN based approaches have been proposed in the literature. One of the most widely used techniques is $\textit{social pooling}$ suggested by Alahi et al.~\cite{alahi2}. The hidden state vectors of the neighboring agents in the same area are first averaged and then fed into the RNN for the target agent together with the current position of the agent. As a result, the RNN learns how to determine the next position of the target agent from the current position, taking into account the motion of the neighboring agents. Motivated by the social pooling method, many similar techniques have been proposed. Xue et al.~\cite{Xue} proposed using the circular grid map instead of the rectangular grid map when calculating pooling vectors since the circle represents social influences among the agents better than the rectangular. Gupta et al.~\cite{Gupta} proposed considering all the agents in a scene and using both the relative positions and the hidden state vectors for the pooling vector calculation. Sadeghian et al.~\cite{SadeghianA} and Xu et al.~\cite{Xu}  used the hidden state vectors of all the agents in order to create a context vector for their social attention mechanism.

\textbf{Agent-space interactions} Existing approaches within this category are generally considering how to use features extracted from the static scene images or spatial positions/routes of interests in a scene. The earlier works usually used the hand-crafted features for extracting meaningful information from the scene images while the recent works let deep convolutional neural networks (CNNs) learn to extract the information. Morris et al.~\cite{Morris}, Ballan et al.~\cite{Ballan}, Katani et al.~\cite{Kitani}, and Walker et al.~\cite{Walker} proposed learning spatial positions/routes of interests in a scene from the scene images or the observed trajectories and using them for the trajectory prediction and activity forecasting tasks. Lee et al.~\cite{Lee} let their RNNs use features from the static scene images to refine the predicted multiple trajectories. Xue et al.~\cite{Xue} introduced a separate RNN to learn possible motion patterns from the static scene images. Sadeghian et al.~\cite{Sadeghian, SadeghianA} used features from the static scene images for their attention mechanism. 

\textbf{Regularization of RNN} Most widely used techniques for regularizing RNN are drop-out~\cite{Srivastava} and $\ell$2-regularization~\cite{Ng} on the trainable parameters of the networks. Recently, Chen et al.~\cite{Chen} proposed an RNN regularization technique taking into account caption generation task. They introduced an additional RNN, called $\textit{auto-reconstructor network}$ (ARNet), which encourages the current hidden state vector of the decoder RNN to embed more information from the previous one. Introducing a discriminator network to generate multiple realistic future trajectories~\cite{Gupta, SadeghianA} can be considered as the regularization on RNN. In this paper, we propose a regularization method based on inverse reinforcement learning framework. The proposed regularization approach differs from~\cite{Gupta, SadeghianA, Chen} in the following points. First, only the encoder RNN is regularized by a reward function so that the encoder RNN learns how to efficiently encode meaningful features extracted from inputs. Second, the reward function evaluates if the predicted positions are realistic along with the scene context information caused by the positions. In contrast, the discriminator networks in ~\cite{Gupta, SadeghianA} use only the past and predicted future trajectories for the evaluation.

\textbf{Inverse reinforcement learning (IRL)} IRL framework also has been adopted for the prediction task since predicting the future from the past can be modeled as Markov decision process (MDP)~\cite{Kitani, Walker, Lee, Kretzschmar}. However, none of them used IRL framework for the regularization purpose. In this paper, we introduce maximum margin IRL framework~\cite{Abbeel} into an encoder-decoder RNN based future trajectory prediction in order to make the networks better utilize the scene context information.
\begin{figure*}
\centerline {\includegraphics[width=16.0cm]{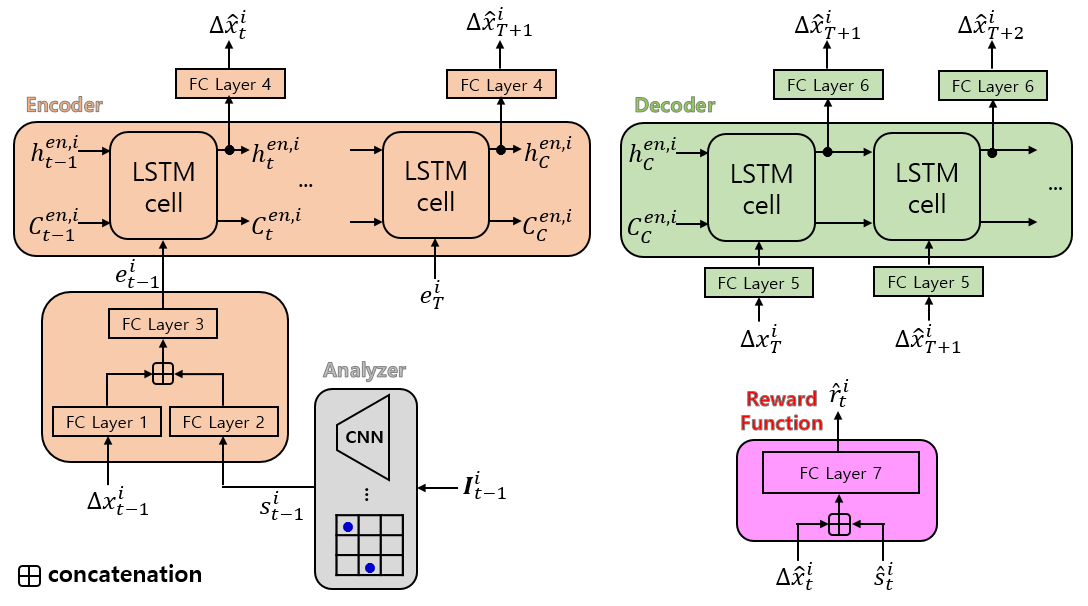}}
\caption{The overall architecture of the proposed model.}
\label{fig:two}
\end{figure*}

\section{Method}
\subsection{Problem Definition}
Let $X^{i}=\{x_{t}^{i}\}_{t=1}^{T}$ and $Y^{i}=\{x_{t}^{i}\}_{t=T+1}^{M}$, respectively, denote the past and future trajectory of the $i$th agent in a scene, where $x_{t}^{i}~(\in\mathbf{R}^{2}$) is the 2D position of the agent at time index $t$. Also, let $S^{i}=\{s_{t}^{i}\}_{t=1}^{T}$ denote the set of the scene context vectors for the $i$th agent, where the scene context vector $s_{t}^{i}$ is calculated from the scene context information $\mathcal{I}_{t}^{i}$. Assuming that there are $N_{a}$ agents in a scene, our target is to predict the future trajectories of all the agents, $\{Y^{i}\}_{i=1}^{N_{a}}$, in the scene given the past trajectories $\{X^{i}\}_{i=1}^{N_{a}}$ and the scene context vectors $\{S^{i}\}_{i=1}^{N_{a}}$.

\subsection{Trajectory Prediction Module}
The proposed model consists of two modules, trajectory prediction module (TPM) and reward function (RF). Figure~\ref{fig:two} shows the overall architecture of the proposed model. TPM, which consists of $\textit{encoder}$, $\textit{decoder}$, and $\textit{analyzer}$, predicts the future trajectories from the past trajectories and the corresponding scene context information. We design TPM to produce the future offsets $\{\Delta x_{t}^{i} = x_{t}^{i} - x_{t-1}^{i}\}_{t=T+1}^{M}$ from the past $\{\Delta x_{t}^{i}\}_{t=1}^{T}$ instead of directly predicting the positions following the suggestion by~\cite{Xu}. For the sake of simple and clear explanation, we will call the offset the position in the rest of this paper. In addition, we will focus on the prediction for the $i$th agent. 

For the future trajectory prediction, the past position $\Delta x_{t-1}^{i}$ and the corresponding scene context vector $s_{t-1}^{i}$ first pass through the fully-connected layers as shown in Figure~\ref{fig:two}. The first two fully-connected layers play the role of the embedding operation. The last one mixes information from the position and the scene context and finally produces the embedding vector $e_{t-1}^{i}$. Note that ReLU operation is applied to all the results from the fully-connected layers in this stage. The scene context vectors are extracted from the scene context information such as the static scene images and the hidden state vectors of the neighbors by the analyzer. The encoder RNN produces the current hidden state $h_{t}^{en,i}$ and cell memory $C_{t}^{en,i}$ from  $h_{t-1}^{en,i}$, $C_{t-1}^{en,i}$, and $e_{t-1}^{i}$.

For the regularization purpose, we let the encoder RNN to predict the current position from the current hidden state vector as follows.
\begin{gather}
\Delta \hat{x}_{t}^{i} = \phi(h_{t}^{en,i}; W_{f4}),~t < T+1,
\label{eqn:one}
\end{gather}
where $\phi(a;W)=b$ is a fully-connected layer, which maps a vector $a$ to a vector $b$ using the parameters $W$. $W_{fP}$ denotes the parameters for the $P$th fully-connected layer, which is described as $\textit{FC layer P}$ in Figure~\ref{fig:two}. The positions predicted by the encoder RNN are evaluated by RF along with the scene context vectors caused by the positions.

The hidden state $h_{t}^{de,i}$ and cell memory $C_{t}^{de,i}$ of the decoder RNN are initialized as $h_{0}^{de,i}=h_{C}^{en,i}$ and $C_{0}^{de,i}=C_{C}^{en,i}$, respectively. With the the last position $\Delta x_{T}^{i}$ in the past trajectory, $h_{0}^{de,i}$, and $C_{0}^{de,i}$, the future positions are predicted as follows.
\begin{gather}
e_{0}^{i} = ReLU(\phi(\Delta x_{T}^{i}; W_{f5})), \label{eqn:two}\\
h_{k}^{de,i} = LSTM_{dec}(h_{k-1}^{de,i}, C_{k-1}^{de,i}, e_{k-1}^{i}; W_{D}), \label{eqn:three} \\
\Delta \hat{x}_{T+k}^{i} = \phi(h_{k}^{de,i}; W_{f6}), \label{eqn:four}\\
e_{k}^{i} = ReLU(\phi(\Delta \hat{x}_{T+k}^{i}; W_{f5})) \label{eqn:five},
\end{gather}
where $1 \leq k \leq M$. $ReLU(.)$ in~(\ref{eqn:two}) and $W_{D}$ in~(\ref{eqn:three}) denote ReLU operation and the parameters for the decoder RNN, respectively.

\subsection{Reward Function}
Let $\hat{s}_{t}^{i}$ denote the scene context vector caused by the position $\Delta \hat{x}_{t}^{i}$. For example, $\hat{s}_{t}^{i}$ can be a pooling vector, which is obtained by the hidden state vectors of the neighbors and the occupancy grid map at the position $\hat{x}_{t}^{i}$. Note that if $\Delta \hat{x}_{t}^{i} = \Delta x_{t}^{i}$ is satisfied, $\hat{s}_{t}^{i} = s_{t}^{i}$ always holds. RF takes as input the state vector $\hat{\mathbf{s}}_{t}^{i} = [\Delta \hat{x}_{t}^{i}$, $\hat{s}_{t}^{i}]$ and produces a reward value $\hat{r}_{t}^{i} \in [0, 1]$ as follows.  
\begin{gather}
\hat{r}_{t}^{i} = Sig(\phi(\hat{\mathbf{s}}_{t}^{i}; W_{f7})), \label{eqn:six}
\end{gather}
where $\textit{Sig}$(.) denotes Sigmoid function. RF is trained to produce a high reward value if $\hat{\mathbf{s}}_{t}^{i}$ is the closest to the ground-truth. Note that the analyzer is utilized by not only TPM but also RF. In addition, the parameters of the analyzer are simultaneously trained via two loss functions, one for TPM and the other for RF as will be explained in the later section. Therefore, the proposed reward function is not a simple FC layer but is the analyzer followed by the FC layer.

\subsection{Simultaneous Training}
TPM is trained to not only produce the future trajectory but also maximize the reward from RF. The loss function for TPM is defined as follows.
\begin{multline}
\mathcal{L}_{TPM}^{i}(W_{TPM}|\textit{W}_{RF}) = \underbrace{\frac{1}{M}\sum_{t=T+1}^{M} ||\Delta x_{t}^{i} - \Delta \hat{x}_{t}^{i} ||^{2}}_{MSE~loss} \\
+ \gamma \underbrace{\mathbf{log}(\frac{1}{T-1} \sum_{t=2}^{T} (r_{t}^{i} - \hat{r}_{t}^{i}) + 1.0)}_{Reward-margin~ loss}, \label{eqn:seven}
\end{multline}
where $r_{t}^{i}$ is the reward value when the ground-truth state $\mathbf{s}_{t}^{i}$ is an input and $\gamma$ is a constant. $W_{TPM}$ and $W_{RF}$ denote the set of the parameters respectively for TPM and RF. The first term of the right hand side of~(\ref{eqn:seven}) is the mean-squared error (MSE) between the ground-truth future position and the future position predicted by TPM. The second term, which we call reward-margin loss, is the margin between the rewards from the ground-truth state vector and its prediction. 

The loss for RF is designed to maximize the margin as follows.
\begin{multline}
\mathcal{L}_{RF}^{i}(W_{RF}|\textit{W}_{TPM}) = - \mathbf{log}(\frac{1}{T-1} \sum_{t=2}^{T} (r_{t}^{i} - \hat{r}_{t}^{i}) + 1.0). \label{eqn:eight}
\end{multline}
The simultaneous minimization of the losses (\ref{eqn:seven}) and (\ref{eqn:eight}) can be regarded as a special case of the iterative algorithm presented by~\cite{Abbeel}, where the encoder plays the role of a policy to be found. The proposed simultaneous training can be summarized as Algorithm 1.
\begin{algorithm}[t]
\caption{}
\begin{algorithmic}[1]
\STATE \textbf{input:} $\{X^{i} \}_{i}$, $\{Y^{i} \}_{i}$, $\{\mathcal{I}_{t}^{i}\}_{i,t}$, $W_{TPM}$, $W_{RF}$
\STATE \textbf{output:} Updated $W_{TPM}$ and $W_{RF}$
\STATE \textbf{initialization:} $\mathcal{L}_{TPM}=\mathcal{L}_{RF}=0$, $i=1$
\WHILE{$i < N_{a}+1$}
\STATE $\% \textit{~Past traj. pred. and scene context vector calc.}$
\STATE Obtain $\{s_{t}^{i} \}_{t=1}^{T}$ from $\{X^{i} \}_{i=1}^{N_{a}}$ and $\{\mathcal{I}_{t}^{i}\}_{t=1}^{T}$;
\STATE Let the encoder RNN produce $\{\Delta \hat{x}_{t}^{i} \}_{t=2}^{T}$ from $\{\Delta x_{t}^{i} \}_{t=1}^{T}$ and $\{s_{t}^{i} \}_{t=1}^{T}$;
\STATE Update $\{\mathcal{I}_{t}^{i}\}_{t=2}^{T}$ according to $\{\Delta \hat{x}_{t}^{i} \}_{t=2}^{T}$;
\STATE Obtain $\{\hat{s}_{t}^{i} \}_{t=2}^{T}$ from $\{\Delta \hat{x}_{t}^{i} \}_{t=2}^{T}$, $\{X^{j} \}_{j=1, i \neq j}^{N_{a}}$, and the updated  $\{\mathcal{I}_{t}^{i}\}_{t=2}^{T}$;
\STATE
\STATE $\% \textit{~Future traj. pred.}$
\STATE Let TPM produce $\{\Delta \hat{x}_{t}^{i} \}_{t=T+1}^{M}$ from $\{\Delta x_{t}^{i} \}_{t=1}^{T}$ and $\{s_{t}^{i} \}_{t=1}^{T}$;
\STATE
\STATE $\% \textit{~Loss calculation by (7) and (8)}$
\STATE $\mathcal{L}_{TPM} \leftarrow \mathcal{L}_{TPM} + \mathcal{L}_{TPM}^{i}$;
\STATE $\mathcal{L}_{RF} \leftarrow \mathcal{L}_{RF} + \mathcal{L}_{RF}^{i}$;
\STATE $i \leftarrow i + 1$
\ENDWHILE
\STATE \textbf{update:} 
\STATE ~~~~$W_{TPM} \leftarrow W_{TPM} + \alpha \frac{\sigma \mathcal{L}_{TPM}}{\sigma W_{TPM}}$; 
\STATE ~~~~$W_{RF} \leftarrow W_{RF} + \alpha \frac{\sigma \mathcal{L}_{RF}}{\sigma W_{RF}}$;
\end{algorithmic}
\end{algorithm}

\section{Experiments}
\subsection{Datasets}
\textbf{Standford drone dataset (SDD)~\cite{Robicquet}} To validate the proposed model on the dataset where the agent-space interaction has a great influence on the agents' movement, we used SDD. SDD includes 8 unique scenes and each scene has more than 4 videos. Each video has its reference top-view image as well as the trajectories of the agents (e.g., pedestrians, cars) in the video. We selected 20 videos for our experiments (at least one video from each scene) since the size of the dataset is too large. To train and test the proposed model, we split the trajectories into segments of 8 seconds (20 frames) each and finally obtained about 200K trajectory segments. We let the proposed model observe 8 frames (3.2 sec) and predict the next 12 frames (4.8 sec) through our experiments. The scene context information was obtained from the reference images and a shallow CNN was used as the analyzer. Note that we scaled the reference images by the factor of 0.25 for the experiments. Finally, we run a randomized 5-fold cross validation without overlapping videos. For example, randomly selected 16 videos were used for training and the remaining four were used for testing. 
\begin{figure}
\centerline {\includegraphics[width=5.0cm]{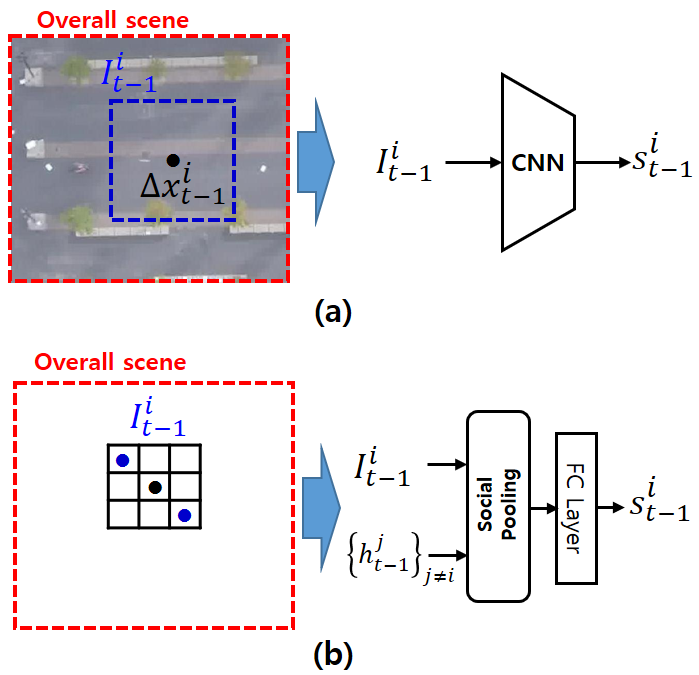}}
\caption{The calculation of the scene context vectors from (a) the static scene context information (a top-view image) and (b) the dynamic scene context information (an occupancy grid map). The black and blue dots, respectively, represent the positions of the target and neighboring agents in a scene.}
\label{fig:three}
\end{figure}

\textbf{Crowd dataset} To validate the proposed model on the dataset where the agent-agent interaction has a great influence on the agents' movement, we used ETH~\cite{Pellegrini} and UCY~\cite{Lerner} datasets, which we call $\textit{crowd dataset}$. ETH and UCY, respectively, contain two (ETH and Hotel) and three (ZARA1, ZARA2, and UCY) videos, which have approximately 1500 pedestrians in total. There are many challenging movement patterns in the trajectories such as collision avoidance, group forming and dispersing. To train and test the proposed model, we split the trajectories into segments of 8 seconds (20 frames) as mentioned above and let the proposed model observe 8 frames (3.2 sec) and predict the next 12 frames (4.8 sec) through our experiments. We used the rectangular occupancy grid map as the scene context information and let the pooling operation~\cite{alahi2} followed by a fully-connected layer be the analyzer. Finally, the leave-one-out approach where 4 videos are used for training while the remaining one is used for testing was taken.

\subsection{Baselines and Evaluation Metrics}
We compare the proposed model against the following baselines: 

\indent $\bullet~\textit{LinearReg}$~: A linear regressor that estimates model parameters by minimizing the least square error.

\indent $\bullet~\textit{S-LSTM}$~: Social LSTM proposed by Alahi et al.~\cite{alahi2}.

\indent $\bullet~\textit{DESIRE-SI-IT0 Best}$~: A deep generative model proposed by Lee et al.~\cite{Lee}. The performance of the model DESIRE-SI-IT0 Best with top 1 sample is reported.

\indent $\bullet~\textit{SGAN-kV-n}$~: Social GAN proposed by Gupta et al.~\cite{Gupta} where $k$ signifies the number of trajectory samples for their variety loss calculation. $n$ denote the number of the candidate future trajectories produced by the network from one past trajectory.

\indent $\bullet~\textit{CarNet}$~: Clairvoyant attentive recurrent network proposed by Sadeghian et al.~\cite{Sadeghian}.

We refer the proposed model without and with IRL framework as $\textit{Ours}$ and $\textit{Ours-IRL}$, respectively. In addition, the proposed model with ARNet~\cite{Chen} is referred as $\textit{Ours-AR}$. The results of $\textit{Ours-AR}$ presented in the following sections were obtained as follows. The parameters of $\textit{Ours-AR}$ were first initialized by the trained parameters of $\textit{Ours}$. Next, ARNet and TPM were simultaneously trained via the loss function proposed in~\cite{Chen}. The parameter $\lambda$, which balances the contributions from TPM and ARNet to the loss function, was experimentally chosen for each test video. For more details, see section 4.2 in~\cite{Chen}.

For the evaluation metrics, we used average displacement error (ADE) and final displacement error (FDE) following the previous studies~\cite{alahi2, Gupta, Sadeghian}.
\begin{table*}
\begin{center}
\begin{tabular}{|c|c|c|c|c|c|c|c|c|c|}
\hline
$\textbf{Metric}$ & $\textbf{Dataset}$ & $\textit{LinearReg}$ & $\textit{S-LSTM}$ & $\textit{SGAN}$ & $\textit{SGAN}$ & $\textit{Ours}$ & $\textit{Ours-AR}$ & $\textit{Ours-IR}~(\gamma=0.1)$ \\
 &  &  &  & $\textit{-1V-1}$ & $\textit{-20V-20}$ &  &  & / $\textit{Ours-IR}^{*}(\gamma)$ \\
\hline\hline
\multirow{6}{*}{$\textbf{ADE}$} & $\textbf{ETH}$ & 1.31 & 1.09 & 1.13 & $\mathbf{0.81}$ & 0.97 & 0.96 & 0.98 / 0.97 ($10^{0}$)\\
& $\textbf{HOTEL}$ & $\mathbf{0.35}$ & 0.79 & 1.01 & 0.72 & 0.56 & 0.56 & 0.51 / 0.51 ($10^{-1}$)\\
& $\textbf{UNIV}$ & 1.00 & 0.67 & 0.60 & 0.60 & 0.59 & 0.57 & 0.56 / $\mathbf{0.54}$  ($10^{-4}$)\\
& $\textbf{ZARA1}$ & 0.90 & 0.47 & 0.42 & $\mathbf{0.34}$ & 0.43 & 0.41 & 0.42 / 0.41  ($10^{-2}$)\\
& $\textbf{ZARA2}$ & 0.57 & 0.56 & 0.52 & 0.42 & 0.38 & 0.34 & 0.32 / $\mathbf{0.31}$  ($10^{-2}$)\\
\hline
& $\textbf{AVG.}$ & 0.82 & 0.71 & 0.73 & 0.57 & 0.58 & 0.56 & 0.56 / $\mathbf{0.54}$  \\
\hline
\multirow{6}{*}{$\textbf{FDE}$} & $\textbf{ETH}$ & 3.00 & 2.35 & 2.21 & $\mathbf{1.52}$ & 1.96 & 1.97 & 1.99 / 1.95  ($10^{0}$)\\
& $\textbf{HOTEL}$ & $\mathbf{0.81}$ & 1.76 & 2.18 & 1.61 & 1.16 & 1.14 & 1.05 / 1.05  ($10^{-1}$)\\
& $\textbf{UNIV}$ & 2.62 & 1.40 & 1.28 & 1.26 & 1.24 & 1.21 & 1.19 / $\mathbf{1.17}$  ($10^{-4}$)\\
& $\textbf{ZARA1}$ & 2.39 & 1.00 & 0.91 & $\mathbf{0.69}$ & 0.91 & 0.90 & 0.89 / 0.78  ($10^{-2}$)\\
& $\textbf{ZARA2}$ & 1.52 & 1.17 & 1.11 & 0.84 & 0.78 & 0.72 & 0.68 / $\mathbf{0.68}$  ($10^{-2}$)\\
\hline
& $\textbf{AVG.}$ & 2.06 & 1.53 & 1.53 & 1.18 & 1.21 & 1.18 & 1.16 / $\mathbf{1.12}$  \\
\hline
\end{tabular}
\end{center}
\caption{The prediction performances on crowd dataset. Each method observed 8 frames (3.2sec) and predicted the next 12 frames (4.8sec). The unit of ADE/FDE is the meter. The values for $\textit{S-LSTM}$, $\textit{SGAN-1V-1}$, and $\textit{SGAN-20V-20}$ are the same as those reported in~\cite{Gupta}.}
\label{table:one}
\end{table*}

\begin{table}
\begin{center}
\begin{tabular}{|c|c|c|}
\hline
$\textbf{Method}$ & $\textbf{ADE}$ & $\textbf{FDE}$ \\
\hline\hline
$\textit{LinearReg}$ & 32.18 & 75.45 \\
$\textit{S-LSTM}$ & 31.19 & 56.97 \\
$\textit{DESIRE-SI-IT0 Best}$ & 35.73 & 63.35\\
$\textit{CarNet}$ & 25.72 & 51.80\\
\hline
$\textit{Ours}$ & 27.96 & 57.97 \\
$\textit{Ours-AR}$ & 27.96 & 57.72\\
$\textit{Ours-IRL}$ ($\gamma$=1.0)  & 23.13 & 48.28\\
$\textit{Ours-IRL}^{*}$ & $\mathbf{21.54}$ & $\mathbf{44.84}$\\
\hline
\end{tabular}
\end{center}
\caption{The prediction performances on SDD. Each method observed 8 frames (3.2sec) and predicted the next 12 frames (4.8sec). The unit of ADE/FDE is the pixel. The values for $\textit{S-LSTM}$, $\textit{DESIRE-SI-IT0 Best}$, and $\textit{CarNet}$ are the same as those reported in~\cite{Sadeghian}. The values for $\textit{Ours}$, $\textit{Ours-AR}$, $\textit{Ours-IRL}$, and $\textit{Ours-IRL}^{*}$ were obtained by averaging the metrics from the five experiments. For $\textit{Ours-IRL}^{*}$, we set $\gamma$ to $10^{-4}$, $5 \times 10^{-2}$, $10^{0}$, $10^{-4}$, and $10^{0}$, respectively for the five experiments.}
\label{table:two}
\end{table}

\subsection{Implementation Details}
We trained TPM and RF based on Algorithm 1 with Adam optimizer~\cite{Kingma}, using a learning rate of 0.0001. The models were trained for 300 epochs and early-stopping technique was applied to avoid over-fitting. $\ell$2-regularization and drop-out technique were also applied to the parameters of the networks with the regularization weight $0.0001$ and keep probability $0.8$. $\gamma$ in (\ref{eqn:seven}) was chosen experimentally. The hidden dimensions of the encoder and decoder RNNs are the same and were set to 256 and 64, respectively for SDD and crowd dataset. 

As we mentioned above, we used a shallow CNN and the pooling operation~\cite{alahi2} followed by a fully-connected layer as the analyzers respectively for SDD and crowd dataset. Figure~\ref{fig:three} shows how the scene context vectors are obtained from the scene context information through the analyzers. Let $W_{A}$ be the parameters of the analyzers. During the simultaneous training, we let $W_{A}$ be included both in $W_{TPM}$ and $W_{RF}$ so the analyzers learn how to extract meaningful features from the scene context information. More details can be found in the supplementary material. In addition, the trained network parameters and the implementation codes will be made publicly available.

\subsection{Evaluation and Comparison}
Table~\ref{table:one} and~\ref{table:two} show the ADE and FDE performances of the models and baselines respectively for crowd dataset and SDD. Note that we let $\textit{Ours-IRL}$ and $\textit{Ours-IRL}^{*}$ respectively denote the proposed model with fixed and variable $\gamma$. We can see in the table that $\textit{Ours-IRL}^{*}$ shows the best performances among the models which produce one future trajectory from the past trajectory. $\textit{Ours-IRL}$ also shows good performances comparable to $\textit{Ours-IRL}^{*}$. For some test videos, $\textit{Ours-IRL}^{*}$ shows the performances better than $\textit{SGAN-20V-20}$, which produces 20 candidate future trajectories given the past trajectory. Note that the prediction performances of $\textit{SGAN}$ and $\textit{DESIRE}$ will be improved as the number of the candidate trajectories to be produced is increased~\cite{Gupta, Lee}. Therefore, it may not be fair to directly compare the proposed model with $\textit{SGAN}$ and $\textit{DESIRE}$ since the proposed model produces one future trajectory, which is the most probable one given the past trajectory and the scene context information. Finally, it is found from the tables that the introduction of ARNet improves the prediction performance of TPM. One can consider incorporating ARNet into the proposed model to further improve the prediction performance of TPM as follows. TPM and RF are first trained by the proposed alternation method. Next, TPM and ARNet are simultaneously trained by using the loss function proposed in~\cite{Chen}.

\begin{figure*}
\centerline {\includegraphics[width=16.0cm]{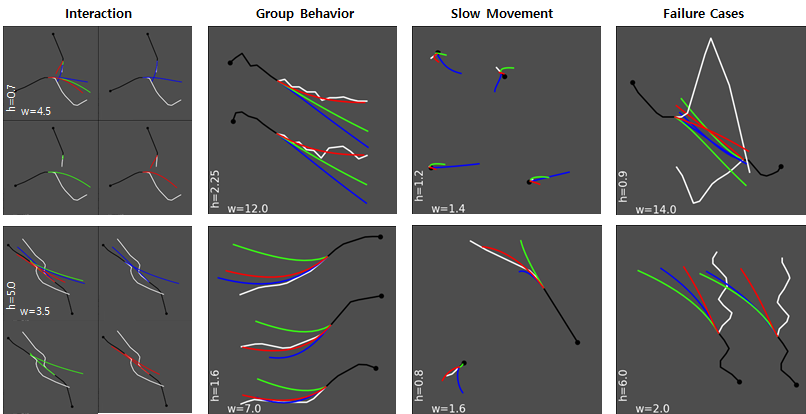}}
\caption{Examples of the trajectory prediction for crowd dataset. Each column shows examples of a different situation. Black and white solid lines, respectively, represent the ground-truth past and future trajectories. Blue, green, and red solid lines represent the trajectories predicted by $\textit{Ours}$, $\textit{Ours-AR}$, and $\textit{Ours-IRL}^{*}$, respectively. The numbers of white color located on the left-bottom area of the images indicate the width and height of the grey region in meters. For visualization purpose, we show the predicted trajectories separately along with the ground-truth trajectory in the first column.}
\label{fig:five}
\end{figure*}

\begin{figure*}
\centerline {\includegraphics[width=16.0cm]{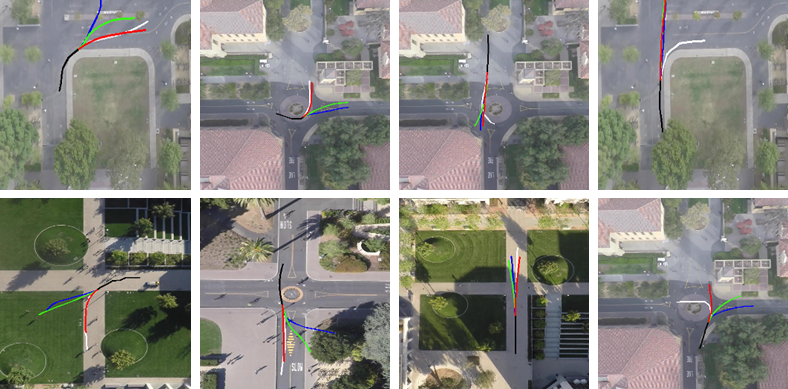}}
\caption{Examples of the trajectory prediction for SDD. Black and white solid lines, respectively, represent the ground-truth past and future trajectories. Blue, green, and red solid lines represent the trajectories predicted by $\textit{Ours}$, $\textit{Ours-AR}$, and $\textit{Ours-IRL}^{*}$, respectively.}
\label{fig:six}
\end{figure*}

\begin{figure*}
\centerline {\includegraphics[width=15cm]{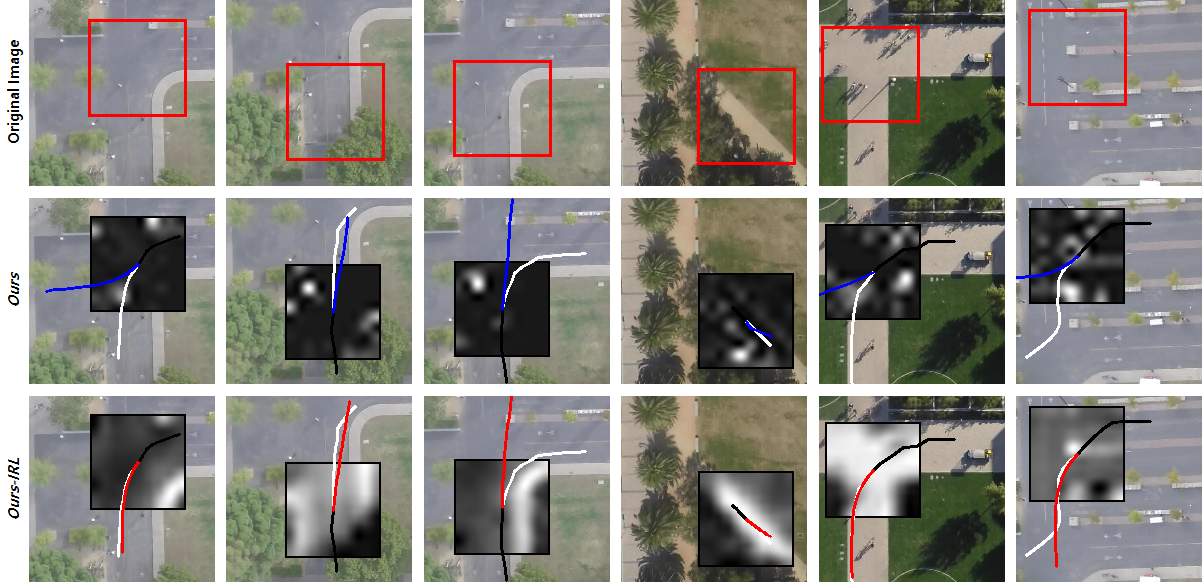}}
\caption{Activation maps obtained from the trained CNNs when the corresponding regions (the red squares in the first row) are used as inputs to the CNNs. Black and white solid lines, respectively, represent the ground-truth past and future trajectories. Blue and red solid lines represent the trajectories predicted by $\textit{Ours}$ and $\textit{Ours-IRL}^{*}$, respectively.}
\label{fig:seven}
\end{figure*}

We show in Figure~\ref{fig:five} and~\ref{fig:six} some examples of the predicted trajectories for crowd dataset and SDD, respectively. The back and white solid lines, respectively, represent the ground-truth past and future trajectories. The blue, green, and red solid lines, respectively, represent the future trajectories predicted by $\textit{Ours}$, $\textit{Ours-AR}$, and $\textit{Ours-IRL}^{*}$. We can see in the figures that $\textit{Ours-IRL}^{*}$ shows the best performance among the three. $\textit{Ours-IRL}^{*}$ predicts the ground-truth trajectories more accurately or produces more plausible trajectories on diverse situations. We also show in Figure~\ref{fig:seven} how the trained CNN interprets the static scene context information. The back squares in the second and third rows represent the activation maps obtained from the trained CNNs when the corresponding regions (the red squares in the first row) are used as inputs. Note that the activation maps were obtained by averaging the absolute of the outputs of the last convolutional layer along the channel axis. The maps in the second and third rows, respectively, correspond to the results of $\textit{Ours}$ and $\textit{Ours-IRL}^{*}$. It is seen in the figure that the shape of the map from $\textit{Ours-IRL}^{*}$ show a stronger correlation with the road structure of the corresponding area than those from $\textit{Ours}$. As a conclusion, the proposed model could learn how to extract meaningful features from the scene context information and encode the features efficiently for the prediction task through our simultaneous training.
\begin{table}
\begin{center}
\begin{tabular}{|c|c|c|c|c|}
\hline
$\textbf{Video}$ & $\gamma$ & lin. traj. & non-lin. traj. & ratio \\
\hline\hline
$\textbf{ETH}$ & $10^{0}$ & 103 & 261 & 0.71\\
$\textbf{HOTEL}$ & $10^{-1}$ & 345 & 852 & 0.71\\
$\textbf{UNIV}$ & $10^{-4}$ & 13541 & 10793 & 0.44\\
$\textbf{ZARA1}$ & $10^{-2}$ & 1159 & 1197 & 0.50\\
$\textbf{ZARA2}$ & $10^{-2}$ & 3575 & 2335 & 0.39\\

\hline
\end{tabular}
\end{center}
\caption{The characteristics of the five videos in crowd dataset. $\textit{lin. traj.}$ and $\textit{non-lin. traj.}$ respectively denote the number of the linear and non-linear trajectory segments in a video.}
\label{table:three}
\end{table}

The last columns of Figure~\ref{fig:five} and~\ref{fig:six} show the cases where the three models failed to predict the ground-truth trajectories. The main reasons are as follows. For crowd dataset, when building the occupancy grid map, only agents at close range were considered so that the three models couldn't consider the motion of the agents far from the target agent. To tackle this problem, one can consider incorporating the pooling method~\cite{Gupta} that takes into account the whole agents in a scene, which may increase the training and prediction times. In addition, some trajectories in the dataset are noisy so the models had difficulty in predicting the future trajectory from the past trajectory and the scene context information. For SDD, disparate trajectories occurring in the same area make the models difficult to predict the ground-truth trajectories. For example, at a crossroad, some may go straight and others turn right. Producing multiple plausible trajectories from one past trajectory can be a good solution to handle the multi-modal nature of the prediction problem. However, selecting the most probable one among the multiple candidate trajectories will not be an easy task.

\subsection{Effect of $\gamma$}
As we mentioned in section 4.4, for the results of $\textit{Ours-IRL}^{*}$, $\gamma$ in (\ref{eqn:seven}) was experimentally chosen for each video. For example, we trained TPM with different $\gamma$ values and chose the one value that shows the best performance on the test video. As seen in the table~\ref{table:one} and~\ref{table:two}, the chosen value is different for each video. This is because the trajectory characteristics of one video can be different from those of other videos. Consequently, the characteristics of the training and the test datasets can be different from each other every time we split the videos in SDD and crowd dataset for the 5-fold cross validation. 

We show in Table~\ref{table:three} the trajectory characteristics of each video in crowd dataset. We can see in the table that each video has a different number of trajectory segments. In addition, the ratio of non-linear trajectory segments in the overall trajectory segments is also different for each video. It seems that the best $\gamma$ has a positive correlation with the ratio. If the ratio is high, $\gamma$ needs to be set to a higher value so that the encoder is trained to pay more attention to the scene context information for the trajectory prediction.

\section{Conclusion}
In this paper, we presented an RNN based prediction model, which predicts the future trajectory from the past trajectory and the scene context information, and its training method based on inverse reinforcement learning framework. In order to make the proposed model better utilize the scene context information for the prediction task, we introduced a reward function, which evaluates the positions predicted by the proposed model along with the scene context information caused by the predicted positions. The reward function was simultaneously trained with the proposed model through the proposed alternation method. Experimental results verified that through the simultaneous training, the proposed model was able to utilize the scene context information much better.



\end{document}